\documentclass[letterpaper, 10 pt, conference]{ieeeconf}
\usepackage{balance}
\usepackage{cite}
\usepackage[pdftex]{graphicx}
\usepackage{amsmath}
\usepackage{algorithm}
\usepackage{algpseudocode}
\usepackage{array}
\usepackage{color}

\author{Ronald Clark$^1$, Sen Wang$^1$, Hongkai Wen$^1$, Niki Trigoni$^1$, Andrew Markham$^1$ % <-this % stops a space
\thanks{*This work was supported by EPSRC.}% <-this % stops a space
\thanks{$^{1}$The authors are with the Department of Computer Science,
        University of Oxford, United Kingdom
        {\tt\small firstname.lastname@cs.ox.ac.uk}}%
}

\begin{document}

\title{Increasing the Efficiency of 6-DoF Visual Localization Using Multi-Modal Sensory Data}

\maketitle
\thispagestyle{empty}
\pagestyle{empty}

\begin{abstract}
Localization is a key requirement for mobile robot autonomy and human-robot interaction. Vision-based localization is accurate and flexible, however, it incurs a high computational burden which limits its application on many resource-constrained platforms. In this paper, we address the problem of performing real-time localization in large-scale 3D point cloud maps of ever-growing size. While most systems using multi-modal information reduce localization time by employing side-channel information in a coarse manner (eg. WiFi for a rough prior position estimate), we propose to inter-weave the map with rich sensory data. This multi-modal approach achieves two key goals simultaneously. First, it enables us to harness additional sensory data to localise against a map covering a vast area in real-time; and secondly, it also allows us to roughly localise devices which are not equipped with a camera. The key to our approach is a localization policy based on a sequential Monte Carlo estimator. The localiser uses this policy to attempt point-matching only in nodes where it is likely to succeed, significantly increasing the efficiency of the localization process. The proposed multi-modal localization system is evaluated extensively in a large museum building. The results show that our multi-modal approach not only increases the localization accuracy but significantly reduces computational time.
\end{abstract}

\section{Introduction}

6-DoF localisation is a key capability required by autonomous humanoid robots and for human-robot interaction (HRI). Consider, for example, the case where a robot needs to locate and pick-up a cup or mug in a domestic workspace. First, the robot needs to find its 6-DoF location and orientation with respect to the stored map. Once this is achieved, the robot can shift its attention to the last known location of the object, check if it is still there, update its knowledge of the object and then proceed with the task as necessary. Another use-case from the HRI side may involve a user, equipped with a pair of smart-glasses, who can motion his head towards locations or objects of interest and mark these for a robot to attend to. This too requires accurate 6-DoF localisation of the user with respect to a global map. 6-DoF pose estimation, however, is computationally expensive which severely limits its application on resource constrained platforms. A key challenge of operating in real-time using 3D pose estimation is thus to narrow the search through the map to make localisation computationally feasible online. Strategies that exploit side-channel information (eg. WiFi, geo-magnetic field distortions) show a marked improvement over exhaustive matching, but due to noisy sensor information there are many false positives and an overly large candidate subset still needs to be searched.

In this paper, we propose an Efficient Multi-modal Localisation (EM-Loc) system, which leverages multiple modalities to make precise vision-based localisation on resource-constrained platforms more feasible. We exploit the side-channel information and along with an estimated trajectory predict which features will be visible. This intelligent guidance is achieved through a sequential Monte-Carlo process which estimates the posterior distribution of the current location over the nodes which ensures the 3D pose estimation is carried out using only the currently visible features. We show how this dramatically reduces computational time for 6-DoF localisation and achieves high accuracy.

\begin{figure}[t!]
\centering
\includegraphics[width=\columnwidth]{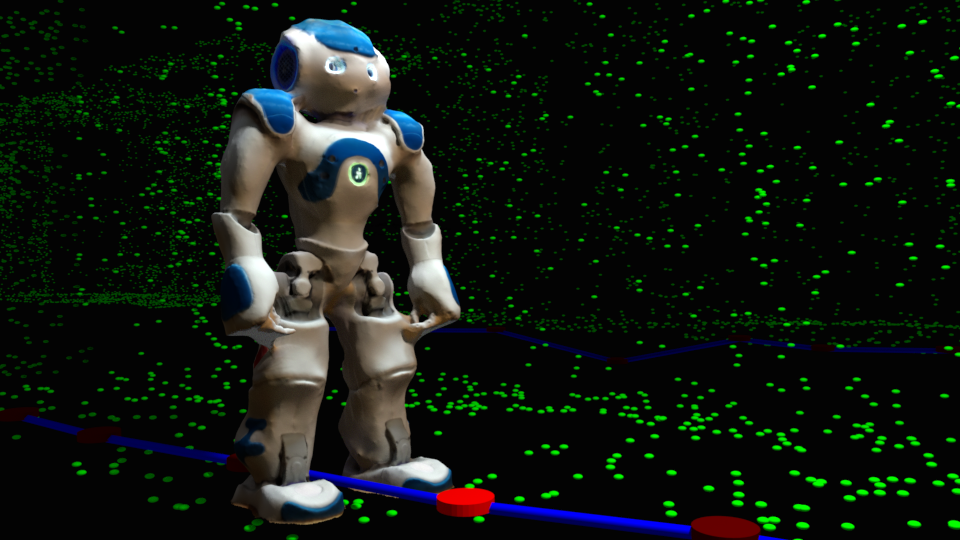}
\caption{localisation against large point clouds comes with a low efficiency and high computational cost making accurate localisation difficult on computationally constrained platforms such as the Nao V5 (with only an Intel Atom @ 1.6 GHz and 48.6 Wh battery). In this paper we reduce the computational load by creating a localisation graph of multiple modalities.}
\label{fig:hero}
\end{figure}

The key contributions of our paper are:

\begin{itemize}
\item We present EM-LOC, an accurate localisation system that can be used across a range of platforms 
\item We propose a localisation policy which exploits side-channel information (WiFi, magnetic) to cut down vision-based localisation time (thus enabling real-time 6-DoF visual localisation) 
\item We evaluate these contributions on data gathered from a large museum building
\end{itemize}

The remainder of the paper is organised as follows: Sec.~\ref{sec:loc} focuses on the two aspects of the EM-LOC system - graph construction and localisation respectively. Sec.~\ref{sec:evaluation} presents an evaluation of the proposed system and Sec.~\ref{sec:conclusion} concludes the paper and highlights ideas for future work. 

%\begin{figure}[h!]
%\centering
%\includegraphics[]{Images/comparison}
%\caption{Difference in the processing pipeline between an appearance-based localisation system, which is common in the mobile computing literature and EM-LOC which provides full 6-DOF visual pose estimates. The image acquisition, and feature extraction processes are the same and incur similar computational costs. In 3D pose computation, however, the feature quantization process is optional as the final pose is computed by accurate 2D-3D feature matching. Our system makes real-time 3D localisation feasible by only attempting to localize where it is likely to succeed. }
%\label{fig:comparison}
%\end{figure}

\section{Related work}
In this section we review existing works related to the EM-LOC system and how they relate to our contribution.
\\
\textbf{Model-based pose estimation:}
When a 3D model of discriminative feature points is available, such as is obtained using Structure from Motion (SfM), then the pose of query images can be found using camera-resectioning. This process first establishes feature matches between the image and the 3D model by either 3D-to-2D \cite{zhang2006image} or 2D-to-3D \cite{sattler2011fast} matching. The camera pose is then found by using RANSAC in combination with a PnP algorithm \cite{lepetit2009epnp}. Camera localisation techniques where direct matching to the 3D model is performed are generally very computationally expensive due to the exhaustive matching that needs to be carried out between the many local feature descriptors. To reduce the matching cost, heuristic techniques have been proposed based on image retrieval methods \cite{irschara2009structure} to find images with 3D points that are likely to be successfully localized against as well as methods that use mutual visibility information to find matches \cite{li2010location}. Although these methods make strides towards reducing the computational burden of large-scale localisation in point clouds, they do not completely solve the problem due to the large decrease in successful localisations (i.e. they trade localisation efficiency for localisation frequency). Our proposed method is orthogonal to these approaches; we use multiple modalities to increase localisation efficiency with negligible decrease in the number of successful localisations.
\\ 
\textbf{Topometric localisation:} The act of localisation does not require exact measurements of a robots pose or position on a map, rather it suffices to estimate the robots location relative to a set of predefined points. These points take the form of nodes in a graph which represent distinct locations or places in the environment and edges which connect neighbouring locations \cite{ulrich2000appearance}. Examples of such systems include CAT-SLAM \cite{maddern2012cat} and the real-time topometric localisation system by Kanade et al. \cite{badino2012real}. The real-time topometric localisation system works in two phases; in the mapping phase the robot drives around the workspace equipped with a GPS and visual sensor. A graph is then created with nodes at fixed interval with each node storing 3D features garnered from the available sensors and stored in a database. In the localisation phase, a Bayesian method is used to estimate the robot with respect to the stored database using the available sensors. Experience based navigation (EBN)~\cite{Churchill2013} extends this formulation by allowing the robot to collect multiple, disconnected segments of the graph during on-line operation. Each segment (called an experience) can be collected during different times of day, weather conditions or seasons and each segment consists of stereo frames linked by edges with odometry information. These segments are then stitched together, allowing the robot to localize under a range of environmental conditions. In this paper, we adopt the graph-based structure of topometric localisation, by using a map of nodes, edges and 3D points. However, we keep fully metric relations between the nodes which allows for accurate localisation. This structure allows us to partition the monolithic model of 3D points, while still being able to index it for fast localisation.
\\
\textbf{Place recognition and appearance-based localisation:}
Appearance-based localisation methods like FAB-MAP \cite{cummins2008fab} and others \cite{Nister2006} recognise places purely based on the appearance of the image. In most cases, the image appearance is described by adopting the bag of visual words (BoVW) model. This process is similar to the large-scale image-retrieval problem. In classical image retrieval, however, the aim is to find as many relevant database images as possible, while appearance-based localisation methods aim at finding co-located images in the database. Thus, approaches such as FAB-MAP are characterised by incredibly high precision with typically low-recall rates. By integrating temporal links and considering places as sequences of images, as is done in SeqSLAM \cite{Milford2012}, the recall rate of appearance-based localisation methods can be significantly improved. Appearance-based methods can also be used for localizing in model-based maps by rendering synthetic views at a large number of candidate locations and orientations. This approach of rendering synthetic views is used, for example, by \cite{irschara2009structure} as a pre-processing step to speed up the direct-matching process in localisation using an SfM model. Because of their efficiency and robustness, appearance-base recognition methods have been popular in mobile phone localisation systems \cite{Zheng:2014:TSI:2639108.2639124} in order to meet the strict resource constraints. Appearance-based methods are also used for loop-closure detections and re-localisations on robotic platforms as they allow one to globally localize against maps of a very large scale \cite{labbe13appearance}. These approaches, however, have the disadvantage of not providing a 6-DoF pose and being far less accurate compared to model-based localisation systems based on local feature point matching. In our system we consider appearance-based features as one of the side-channels (along with WiFi and geo-magnetic measurements), which are used to find model sections against which to localize. Like SeqSLAM, we utilizes temporal/sequential information for global localisation.
\\
\textbf{Multi-modal localisation:} The promise of exploiting multiple modalities for aiding vision-based localisation has been demonstrated in a variety of ways, but not directly for reducing visual processing time. For example, the W-RGBD system \cite{ito2014w} uses WiFi in combination with RGBD images to perform localisation using an architectural floorplan. They model the WiFi signal strength across the floorplan using a Gaussian process and through the use of this model intelligently initialize the particles for Monte Carlo localisation (MCL). They show that this improves the convergence rate and final accuracy of the position estimate. Their goal differs from ours; while they are mainly concerned with using WiFi to increase the convergence rate and global accuracy of MCL we are interested in using the side-channel information to guide the expensive 2D-3D feature matching to increase the efficiency of the RANSAC-based pose computation and thus reduce the computational time required on resource-constrained platforms.

\section{Proposed System}

\begin{figure}[h!]
\centering
\includegraphics[]{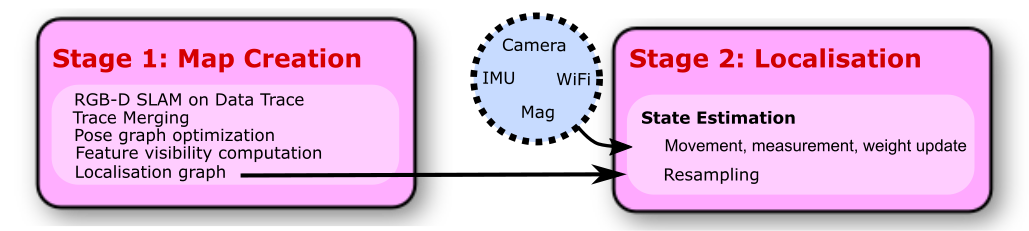}
\caption{EM-LOC consists of a two-tier architecture. The localisation graph is created and updated in the cloud, while the localisation module receives the localisation graph and performs on-line state estimation.}
\label{fig:system}
\end{figure}

Our system consists of a two-tier architecture, as shown in Figure~\ref{fig:system}. The back-end graph assembler is responsible for creating the localisation graph, while the front-end runs on the target platform and performs on-line localisation. The back-end graph assembler receives as input the data collected from mappers during the mapping session of the system which are uploaded to the cloud where the server is implemented. The graph assembler constructs the localization graph by merging data traces from different mappers. The localiser carries out state estimation to position the humanoid with respect to the assembled localisation graph.

\section{Stage 1: Map creation}
\label{sec:assembly}

%\begin{figure}[h!]
%\centering
%\includegraphics[]{Images/segmenting}
%\includegraphics[]{Images/segmenting1}
%\caption{The localisation graph consists of multiple traces which have been merged. Segments are matched in both forward and reverse order to account for paths being traversed in both directions.}
%\end{figure}

We use three types of information in our map creation phase - RGB data from which sparse SIFT features \cite{Lowe2004} are extracted and associated with 3D points (these are needed to perform the 6-DoF localization), depth frames (which are used to determine the visibility of the sparse feature points from nodes) and the multi-modal side-channel information.  Using this data a map of 3D points associated with SURF features and the relative translation between frames is created using an RGB-D SLAM algorithm \cite{endres20143} \footnote{If a depth camera is not available, this process could also be carried out using an image-based Structure-from-Motion process followed by a dense multi-view stereo reconstruction.} The localization graph in our case is a set of nodes, $n_1,n_2,\dots,n_N$ (each frame is considered as a node) which are joined by a set of directed edges $e_{i-1,i}$ that link the nodes. Each edge contains odometry data consisting of the displacement and orientation (computed by the RGB-D SLAM) which describes the metric relation between the nodes. Each edge also contains the side-channel information which includes the distribution of WiFi RSSI values along that edge, geo-magnetic distortion measurements and appearance-based visual features. 

\begin{figure}[t!]
\centering
\includegraphics[width=\columnwidth]{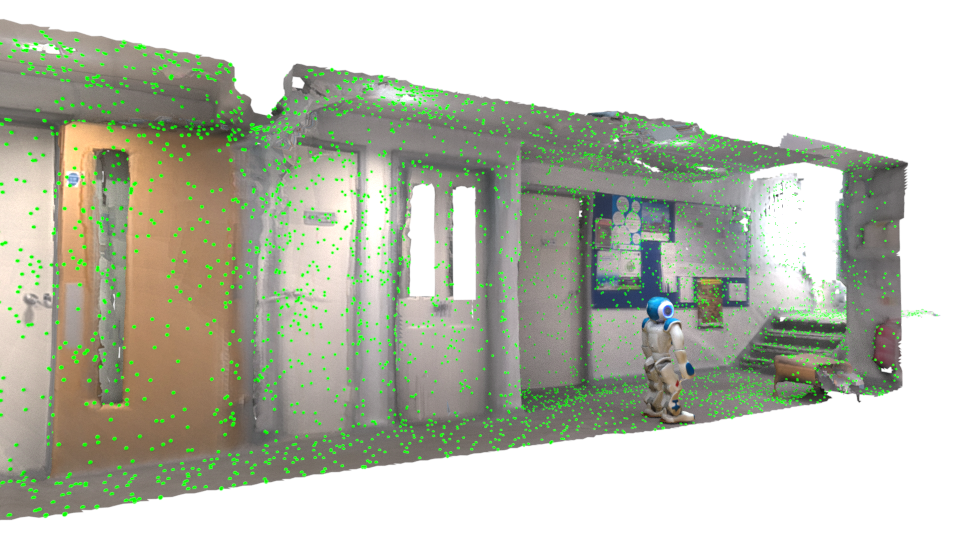}
\includegraphics[]{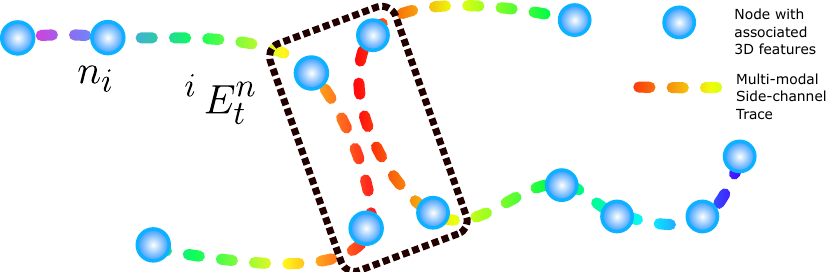}
\caption{Illustration of a 3D model for localization with millions of candidate feature points (green dots).In this paper, we propose EM-Loc (Efficient Multi-modal Localization) that integrates these features in a localization graph and interweaves it with additional (side-channel) sensory data. Our localiser uses this side-channel information to predict which features will give successful visual localization. This cuts down the matching time and enables real-time and fine-grained localization over vast areas.}
\label{fig:example}
\end{figure}

The structure of the side-channel features is as follows: 
\\
\noindent \textbf{WiFi}: The readings are composed of the received signal strength, $s$, for each access point observed in the environment. As described in \cite{shu2015magicol}, the RSS of typical radio-based measurements does not change very rapidly with typical humanoid/human mobility and thus the distribution is collected over a finite duration of time which is used as the feature,
$ {E}^{wifi}_i = \left[\mbox{s}_1,\mbox{s}_2,\cdots \mbox{s}_n \right] $
\\ 
\noindent \textbf{Magnetic}: Similarly, we capture the magnetic field strength data over a short time segment. The magnetic data consists of the magnitude, $m$, of the geo-magnetic field calculated from the 3 components in the $x,y$ and $z$ axes, $ {E}^{mag}_i = \left[m_1,m_2,\dots, m_n \right] $
\\ 
\noindent \textbf{Image}: We further capture images along each experience and use appearance-based features derived from these images as additional side-channel information. The full images themselves are not stored, rather we detect and compute the distinctive visual features which are present in the image. Each feature is then quantized into BovW vector, where the word vocabulary, $w_1,w_2,\cdots,w_n$ has been precomputed using k-means clustering. The final representation of the image consists of a histogram of the word occurrence counts
$ {E}^{img}_i = \left[ h_1,  h_2, \cdots, h_n\right] $
As the quantization incurs additional cost, the inclusion of appearance-based features is entirely optional in our system, and if these features are used the vocabulary is kept very small. 
\\ \\
The graph consists of data traces, which may be collected during different times, allowing the map to be readily extended. Many disjoint data traces may be generated by mappers traversing an area. The localisation graph thus consists of short segments, along with their nodes and edges, which need to be linked together at certain co-location points. The graph assembly process harnesses the side-channel information to establish these co-location links across multiple data traces. Given a trace denoted as ${E}^n_i$ where $n$ is the modality, $i$ is the trace index and $t$ is the timestamp at which the data was recorded, a co-location link established across multiple traces when
\begin{align}
\exists t_2 \mbox{ s.t. } \max P(^i{E}^n_{t_1}|{^iE}^n_{t_2}) > \tau.
\end{align}

This means is that for each data frame of a certain modality that is assigned to a place, some other modality, already belonging to that place shares a strong similarity with its partner in the current trace. After the co-location links have been established the entire localisation graph is optimized using a robust pose graph optimization through the GT-SAM library \cite{dellaert2012factor} with the Vertigo extension \cite{sunderhauf2012switchable}. The graph construction process is carried out off-line and also updated using data collected during additional mapping sessions. The end result of this process is a consistent representation of the world in the form of a linked multi-modal localization graph.  

\begin{figure}[h!]
\label{fig:visibility}
\centering
\includegraphics[width=\columnwidth]{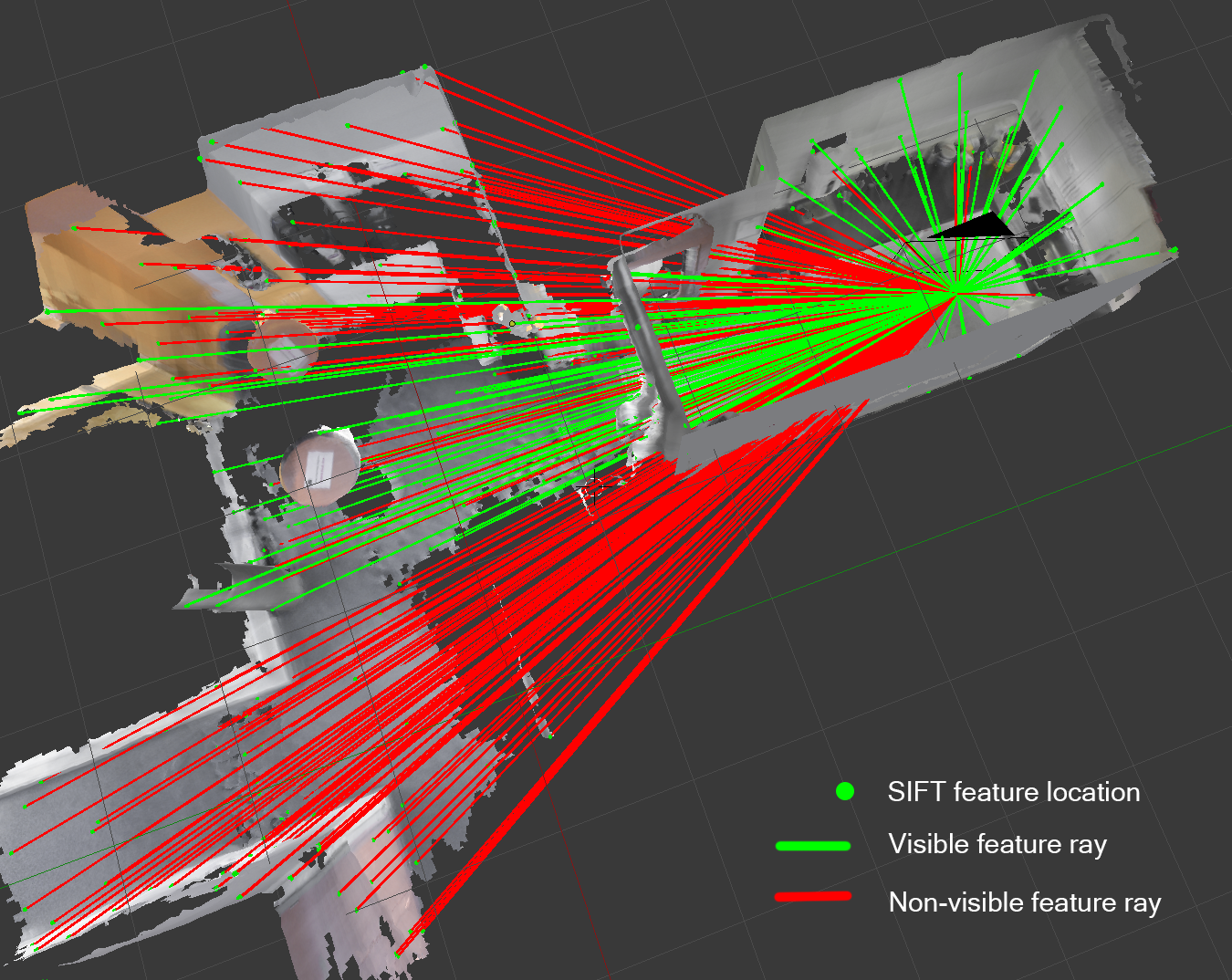}
\caption{In order to determine the visbility of features from each node location, we make use of the depth frames. The visbility is computed by ray-casting from the node center to the stored 3D location of the feature.}
\end{figure}

Once the graph has been created and optimized it consists of a large number of nodes each associated with a small number of feature points which represent those that were visible in the frame during the mapping process. However, once the consistant map has been created, many feature points may be visible from a node which is not taken into account by the initial mapping process. We therefore utilise a post-processing step to determine the true point visibility for each node. This is achieved by raycasting from the node to the 3D location of each SIFT feature stored in the map and checking for ray intersections with depth frames registered to their world locations. If no hit occurs occurs, the SIFT feature is visible from that node and is added to the node's feature-set. This process is illustrated in Fig. \ref{fig:visibility}. Each node is associated with a set of 3D features which comprises those features that are visible from the node's location. 

\section{Stage 2: Localisation}
\label{sec:loc}

In this section we describe our localiser which runs on the humanoid with the objective of providing efficient 6-DoF pose estimates. We adopt a sequential Bayesian approach to perform localisation in the established localisation graph. 
\[ p(n_i,\mathbf{X}_i,\Theta_i|z_{i:t}) = p(\mathbf{X}_i,\Theta_i|n_i,z_{i:t}) p(n_i|z_{i:t}) \]
Where $n_i$ are the nodes of the localisation graph as defined before, $\Theta_i$ is the orientation (yaw, pitch, roll) of the device, and $\mathbf{X}_i$ is the location of the humanoid \emph{with respect to} the current node $n_i$. The pose estimates relative to a node are independent of the node, the state estimation problem reduces to 
\[ p(n_i,\mathbf{X}_i,\Theta_i|z_{i:t}) = p(\mathbf{X}_i,\Theta_i|z_{i:t}) p(n_i|z_{i:t}) \]
\\
In order to perform localisation, we use a particle filtering approach. This enables us to employ a Rao-Blackwellized approach to performing the localisation on the localisation graph where the term $p(n_i|z_{i:t})$ keeps track of the global localisation of the humanoid in the localisation graph, while the term $p(\mathbf{X}_i,\Theta_i|z_{i:t})$ ensures that the local 6D pose computations remain consistent. In order to ensure that the pose computation remains lightweight, we make the assumption that the local pose estimate produces a uni-modal Gaussian. Our particle-filter is initialized and updated as follows.

\subsection{State and initialization:}
The particle structure is split into two components which consists of an estimate of the humanoid's current location in the localisation graph represented by the node $n_i$ and $(x_i,y_i,z_i,\theta_i)$ which is the 6-DoF pose of the humanoid relative to node $n_i$.
\[ \mbox{Particle: } X_i = \left[\{n_i\},\right] \left[ \{\mathbf{\mu}_{\mathbf{X}_i,\Theta_i} \mathbf{\Sigma}_{\mathbf{X}_i,\Theta_i} \} \right]\]
We uniformly initialize the particles over the entire localisation graph.

\subsection{Measurement}
\label{sec:features}
In order to utilise the incoming sensory measurements for localisation in our localisation graph, we need some means of relating the measurements to the data stored in our graph. This is achieved by utilising likelihood functions for each modality as a function of the system state. In particular, we use the following likelihood functions:
\\ 
\noindent \textbf{RSSI:} For RSSI data, the Kullback-Liebler divergence, $D_{KL}$, is used as the metric to compare measurements coming from individual access points.
\[ p({E}^{wifi}_i|E{i}^{wifi}_j) \propto e^{-D_{KL}\left({E}^{wifi}_i, {E}^{wifi}_j)\right)} \]
\\
\noindent \textbf{Magnetic:} The magnetic features consist of a series of geo-magnetic measurements strung together in a series. As the humanoid may have been moving at different speeds between the stored and query feature, we use dynamic time warping (DTW) as a measure of their similarity   
\[ p({E}^{mag}_i|{E}^{mag}_j) \propto e^{-DTW({E}^{mag}_i, {E}^{mag}_j)}. \]
\\
\noindent \textbf{Image:} To compute the likelihood of a new image given one in the stored experience we use use the FAB-MAP approach \cite{cummins2008fab}. FAB-MAP takes as input the image's BoW vector and produces a normalized likelihood measure for each stored location, $p({E}^{img}_i|{E}^{img}_j)$.
\\
We compute the weight by assuming that each of the side-channel likelihoods is independent. The importance weights, $w^i_t$, for the particles are thus calculated as the product of the individual terms

\[ w^i_t = p({E}^{wifi}_i|E{i}^{wifi}_j) p({E}^{mag}_i|{E}^{mag}_j) p({E}^{img}_i|{E}^{img}_j). \]

\subsection{Odometry}
For the process update of our sequential Bayesian filter, we use odometric measurements. Many humanoid odometry estimation schemes rely on incorporating domain constraints for a specific application, along with a complex dynamics model \cite{yamane2012systematic} and state-estimation methods. In this paper, however, we are interested in the case where the IMU is not firmly attached to the body of the subject being tracked with the goal of allowing our approach to be transferred seamlessly from a humanoid to human (eg. human wearing a pair of smart-glasses). In this case such strict and clear-cut assumptions does not exist and thus, rather than relying on a state-space approach and encoder data, most methods extract features from the IMU traces and use these these for odometry estimation in what is know as pedestrian dead reckoning (PDR) \cite{park2011height}. PDR performs 3 main operations; step detection, step-length estimation and heading calculation. By coupling the stride length estimates and step event detections derived from these features, the PDR is able to roughly estimate the relative position of the subject. Step events are detected as maximum of the z-axis accelerometer data and the heading is tracking using the the gyroscope and magnetometer \cite{kleiner2007decentralized}. We use a simple step-frequency which has been shown to work well in practice \cite{renaudin2012step}. This model relates the step length to the step frequency, $l_s = \alpha f + b$, where $f$ is the step frequency and $\alpha$ and $b$ are constants as described in \cite{renaudin2012step}.

\subsection{Pose Estimation using 3D features} 
\label{sec:3dpose}

6-DoF pose estimation using 3D features is performed using the EPnP algorithm \cite{lepetit2009epnp}. Using a set of 2D features detected in an image at locations, $u_i,v_i$ matched to 3D features $x^N_i, y^N_i, z^N_i$ stored relative to the coordinate system of node $n$. The pose, $\mathbf{X}, \mathcal{\theta}$, of the the camera which recorded the image relative to the world coordinate system, is computed using a PnP solver. We use the standard EPnP solver wrapped in a RANSAC loop as implemented in OpenCV \cite{opencv_library}.

\subsection{EKF Process and Measurement Update} 
Unlike the side-channel information which gives inexpensive, but noisy and outlier-prone estimates of the current location, the 6-DoF pose computation gives outlier-free measurements. An internal EKF is used to keep track of the pose mean $\mathbf{\mu}_{\mathbf{X}_i,\Theta_i}$ and the covariance $\mathbf{\Sigma}_{\mathbf{X}_i,\Theta_i}$, \emph{relative} to a particular node. The process update of the EKF is carried out by using odometric measurements.

In particular, process update is carried out as follows
\begin{align*} 
x^t_n &=  x^t_n + l_s \times cos(\theta_\phi) \\ 
y^t_n &=  y^t_n + l_s \times sin(\theta_\phi) 
\end{align*}
Where the step lengths $l_s$ and step detections are obtained from the PDR. For the EKF measurement, a 6-DoF pose computation is done using the 3D features, this yields an observation $z = \hat{x}^N_i, \hat{y}^N_i, \hat{z}^N_i,\Theta_i$ relative to the 3D node, $n_i$, to which those features belong. Using this measurement, the Kalman gain is calculated using the standard gain equations \cite{haykin2001kalman} and the mean and covariance updated using this gain. We concurrently perform the measurement update for the weights of each particle sharing the same node $n_i$, which significantly reduces the computational effort.

\subsection{Update and resampling}
\begin{figure}[t!]
\centering
\includegraphics[width=\columnwidth]{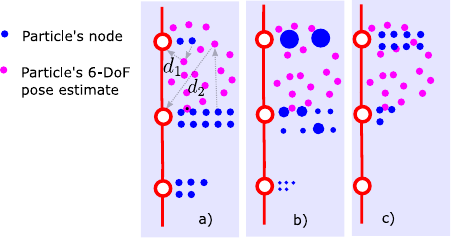}
\caption{a) Before particle re-weighting many particles contain 6-DoF pose estimates which are far from their current graph node. b) After weight modification, particles with with nodes far from their 6-Dof estimates are given a lower weight. c) After re-sampling more particles have closely-related node and 6-DoF pose estimates. }
\label{fig:reproject}
\end{figure}

The weights consider only the likelihoods of the particles at the discretised node-level. However, our particles contain more information i.e. in the form of a continuous 6-DoF pose estimate (maintained by the EKF). As the dead reckoning and 6-DoF pose estimate may lead to the particle deviating from the current nearest node and coming closer to another one, we further adjust the weight of the particles such that those closer to the predicted pose have a greater weight.
\[ w_k^{i'} = w_k^{i'} e^{-d}. \]
The quantity $d= \sqrt{(x_k^{i'})^2+(y_k^{i'})^2+(z_k^{i'})^2}$ represents the distance of the particle's 6-DoF pose estimate to the location its node $i'$. The transformation to a particular node's coordinate system is done using the information stored along the edges in the graph. This weight adjustment is illustrated in Fig. \ref{fig:reproject}.

%\subsection{Additional visual links}
%\label{sec:map}

%\begin{algorithm}
%  \caption{Rewighting using additional visual links \label{alg:selecting_places}}
%  \begin{algorithmic}[1]
%    \Require{Localization graph ${G}$, particle weights $w_i$}
%    \Statex
%    \Function{Update weights}{${G}$,$w_i$}}
%\Comment{$\oplus$: bitwise exclusive-or}
%        \State {Establish auxiliary visual links $e_{aux}$}
%        \For{${n}_i \textrm{ in } {G}$}
%          \While{$\exists \textrm{ sub-chain connected to } {n}_i$}
%            \State $\textrm{$w'_i$ = \alpha $w_i$}$
%            \State $\textrm{Proceed to next sub-chain}$
%          \EndWhile
%        \EndFor
%      \State $\textrm{normalize} $ $w_i$  
%      \State \Return{$w_i$}}
%    \EndFunction
%  \end{algorithmic}
%\end{algorithm}

%\subsection{3D Feature Visibility Computaion}
%\label{sec:visibility}
%\begin{algorithm}
%  \caption{3D Feature Visibility Computaion \label{alg:visibility}}
%  \begin{algorithmic}[1]
%    \Require{Localization graph ${G}$, features $f_i$, depth frames $I_i$}
%    \Statex
%    \Function{Update weights}{${G}$,$w_i$}}
%\Comment{$\oplus$: bitwise exclusive-or}
%        \State {Establish auxiliary visual links $e_{aux}$}
%        \For{${n}_i \textrm{ in } {G}$}
%          \While{$\exists \textrm{ sub-chain connected to } {n}_i$}
%            \State $\textrm{$w'_i$ = \alpha $w_i$}$
%            \State $\textrm{Proceed to next sub-chain}$
%          \EndWhile
%        \EndFor
%      \State $\textrm{normalize} $ $w_i$  
%      \State \Return{$w_i$}}
%    \EndFunction
%  \end{algorithmic}
%\end{algorithm}

\subsection{Localisation policy}
A major bottleneck in using full 3D localization for the purpose of mobile device localisation is the severe limitation which the computational constraints of the mobile device places on the number of 3D features which can be localized against while still allowing for real-time operation. To address this, we use a localisation policy which performs the expensive 2D-3D matching only in nodes where it is essential, guided by the current distribution of particles on the graph. 

For reliable localisation, we would like to ensure that 2 criteria are met; 1) the particle distribution should faithfully represent the posterior distribution of the humanoid's location on the graph and 2) we only want to try to localize on a limited number of nodes and these should have a high probability of being the humanoid's actual location.

In order to satisfy 1) we use a metric based on the importance weights obtained from the previous time instant, $w^i_{t-1}$. These weights are used to calculate the ``effective number of particles''  
$
N_{eff} = \frac{1}{\sum \hat{\omega^i}^2_{t-1}}, 
$
which is a metric that gauges how well the particles represent the posterior \cite{doucet2000rao}. 

The localisation policy then uses this metric to determine when localisation should be attempted. If the effective number of particles is high enough, the localiser attempts to localize in the $k$ most likely nodes. Typically, the 3D matching only has to be done on 1 or 2 nodes and each of these frames \emph{naturally} contains only a fraction of the most relevant 3D features needed for localisation. The full localisation algorithm is detailed in Alg. \ref{alg:policy}.

\begin{algorithm}
  \caption{Localizer \label{alg:policy}}
  \begin{algorithmic}[1]
    \Require{localisation graph ${G}$, Particles $X_i$}
    \Statex
   \Function{Localize}{${G}$,$X_i$}
        \State $\textrm{Calculate } N_{eff} $
       \For{{$m \textrm{ = } {1:k}$}}
          \If{$N_{eff} > \tau_{eff}$}
            \State  $\textrm{Select } m^{th} \textrm{ most common node }$
            \State  $\textrm{Pose computation using } m's \textrm{ 3D features}$
          \EndIf
        \EndFor
      \State \Return{$\mathbf{X}_t,\mathcal{\theta}_t$} \Comment{Proceed with EKF update}
    \EndFunction
  \end{algorithmic}
\end{algorithm}

The quantity $k$ is a very important parameter as it determines the number of nodes in which localisation is attempted. It therefore directly affects the number of feature matches that are carried out during the 6-Dof pose estimate. By adjusting this quantity, the number of feature matching can be controlled as required and thus allows the localisation time to be scaled according to the requirements of the platform.

\section{Experiments}
\label{sec:evaluation}

In order to adaquately evaluate EM-LOC, we gauge its performance along three performance axes. The first is the \textbf{localisation time}. As the 6-DoF localisation is the most computationally expensive component which we seek to minimize, we define as the localisation time as the time it takes from receiving the current visual frame from the camera to the time that the 6-DoF pose for the frame is determined by the RANSAC-based pose computation. The second factor is the \textbf{localisation efficiency}. The localisation efficiency is defined as the number of successfully localised divided by the total number of images for which localisation was attempted. This is important as the RANSAC-based pose computation may not localise a 6-DoF pose if it does not find enough inliers. In our case we use 8 inliers as the threshold for successful localisation. The final aspect we evaluate is the \textbf{localisation accuracy} which is simply the difference between the system's current estimate of the angle and orientation and the ground truth pose. 

%\begin{figure}[h!]
%    \centering
%    \includegraphics[width=0.7\columnwidth,angle=90]{Images/dataset1.pdf}
%    \caption{Overview of the area of the two datasets used in the experiments.}
%    \label{fig:datasets}
%\end{figure}
\begin{figure}[t!]
    \centering
    \includegraphics[width=0.3\columnwidth,angle=90]{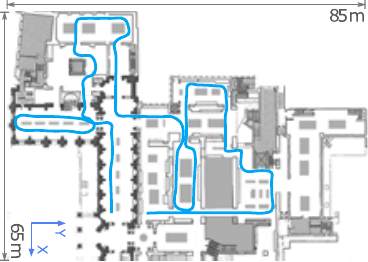}
    \includegraphics[width=0.69\columnwidth]{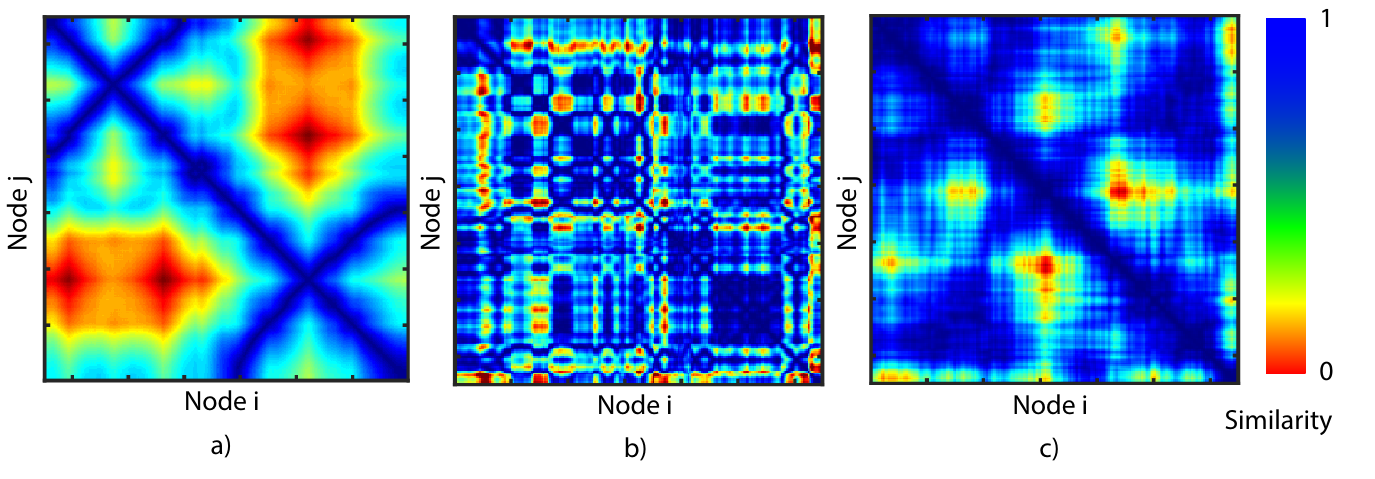}
    \caption{Visualization of the magnetic and WiFi similarities for the data
used in the museum experiment. a) shows the ground-truth metric distance
between the nodes b) shows the magnetic signal similarity and c) the WiFi
similarity}
    \label{fig:similarities}
\end{figure}

For fair comparison, we compare to two existing methods in terms of both accuracy and processing time. The
first system, which we use as a baseline for comparison is the EBN framework \cite{Churchill2013}. The EBN method stores frames along with 3D feature points against which incoming images can be localised. We consider two variants of the framework. The first carries out an exhaustive matching against 3D points belonging to all available nodes stored in the map - in order to vary the number of nodes that are used in the localisation, we randomly select $k$ nodes from the graph (we label this method A1). In the second method for comparison, we make use of the side-channel data to rank the nodes which are selected (A2). The ranking is computed by using the likelihoods of Sec. \ref{sec:features} as features and using the independence assumption - ranking the nodes as the product of the features in the same way as the weights of the particles are calculated in EM-LOC. Our third method for comparison is based on Travi-Navi \cite{zheng2014travi}, which is a state-of-the art teach-and-
repeat pedestrian bi-pedal navigation system from the mobile computing literature. Travi-Navi simply uses an SVM to produce an estimate of the current location from the WiFi, magnetic and appearance-based visual data. For our purposes, we use the SVM to rank the nodes on which the 3D pose estimation is performed (A3).

Our evaluation data was gathered from a
large museum area. The area was chosen for numerous
reasons; firstly, its size $(\approx 5000m^2)$ allows for thorough
evaluation of our approach; secondly it is representative of a
practical implementation setting of our system - a popular
tourist attraction and public site with much commercial
interest for a viable localisation service. Furthermore, the
museum interior is a very complex space - filled with vast
open rooms, narrow corridors and dim-lighting which makes
it a challenge for vision-based localisation methods. In order to evaluate such a large area and considering the locomotion speed of current humanoid robots, the data for our experiments were collected by two humans equipped with a pair of Google Glasses. This is reasonable as our odometry does not rely on modelling intricate dynamics. For offline procedure of map creation, RGB-D frames were collected using a Project Tango device and run through an RGB-D Slam algorithm \cite{endres20143}. The same procedure was used to collect ground-truth locations for the online localisation tests. We stress, however, that all the reported online localisation
results are obtained using the data from Google Glass. We test at a total of 167 test locations across the
museum. Our experiments were run on an Intel Core i5-2467M @ 1.60GHz, which is similar in performance to that of the Nao v5.

\begin{figure}[h!]
    \centering
    \includegraphics[width=0.8\columnwidth]{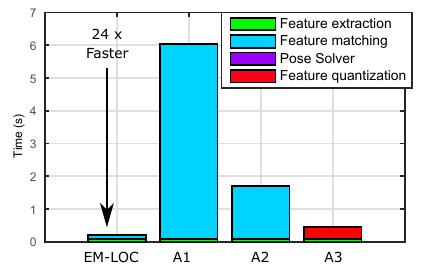}
    \caption{Comparison between the processing times for EM-LOC
system, and comparison methods A1, A2 and A3 (see text)}
    \label{fig:accuracy_timing}
\end{figure}

\begin{figure}[h!]
    \centering
    \includegraphics[width=\columnwidth]{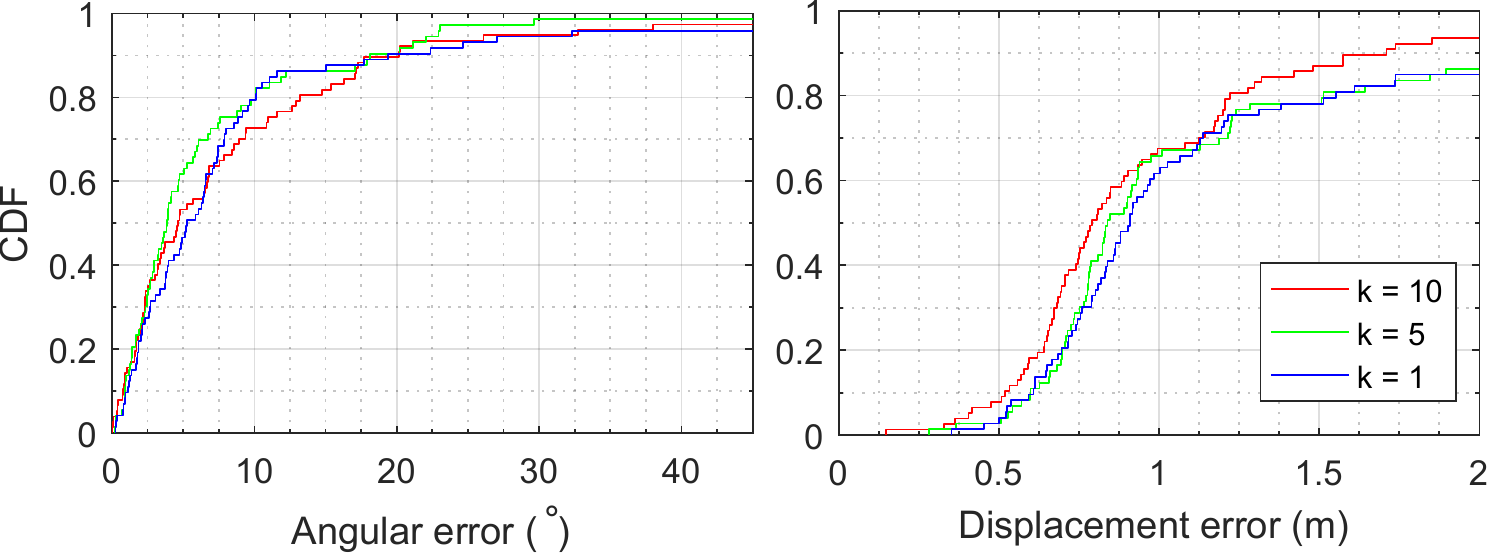}
    \caption{Cdf plot of the orientaation and translation error of EM-LOC from various values of $k$.}
    \label{fig:cdfs}
\end{figure}

\begin{figure}[h!]
    \centering
    \includegraphics[width=\columnwidth]{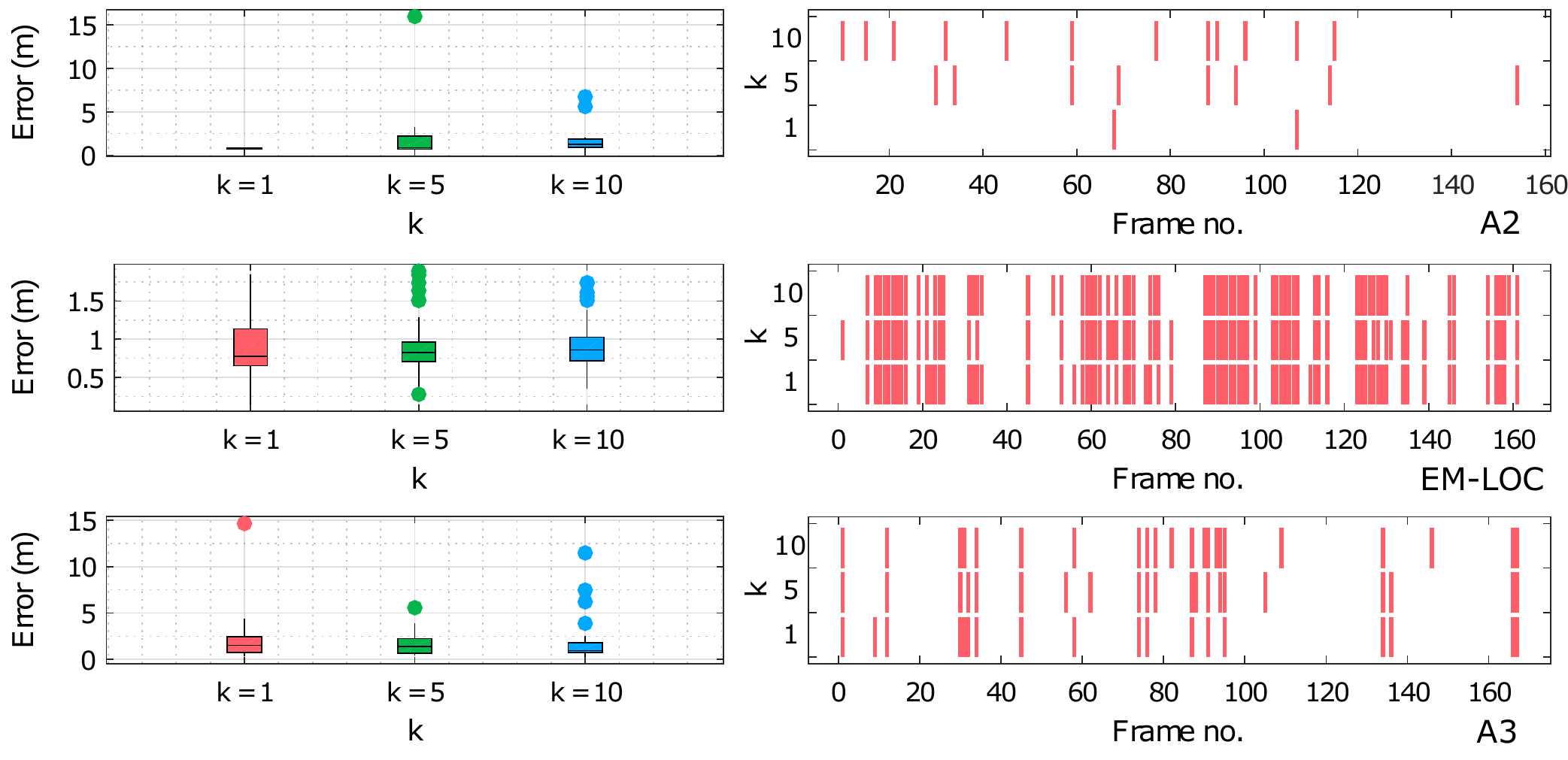}
   \caption{Evaluation of the localisation accuracy and the localisation efficiency. The right subplots show the successful localisations for each of the input test images.}
    \label{fig:evaluation}
\end{figure}

The error distribution for EM-LOC, A2 and A3 at 3 values of $k$ is reported in the left subplots of Fig. \ref{fig:evaluation} (A1 is not plotted in the graph as the random node decimation performs significantly worse than the other methods at these low $k$ values) and the CDF in Fig. \ref{fig:cdfs}. The
EM-LOC system clearly outperform methods A2 and A3
in terms of localisation accuracy. This can be explained through two observations. Firstly, EM-LOC, has the ability to better propose
matching candidates for the localisation. This means that
nodes where localisation may be possible, but of low
accuracy (such as distant nodes) are automatically excluded. Secondly, by leveraging multi-modal data, EM-LOC is able to resolve visual ambiguities such as corridors of similar appearance which are common very common in indoor environments and in many cases cannot be resolved using visual data alone. These two factors contribute to the 6-DoF pose estimation in EM-LOC producing fewer outliers and more accurate pose estimates. 

\begin{figure}[h!]
    \centering
    \includegraphics[width=0.8\columnwidth]{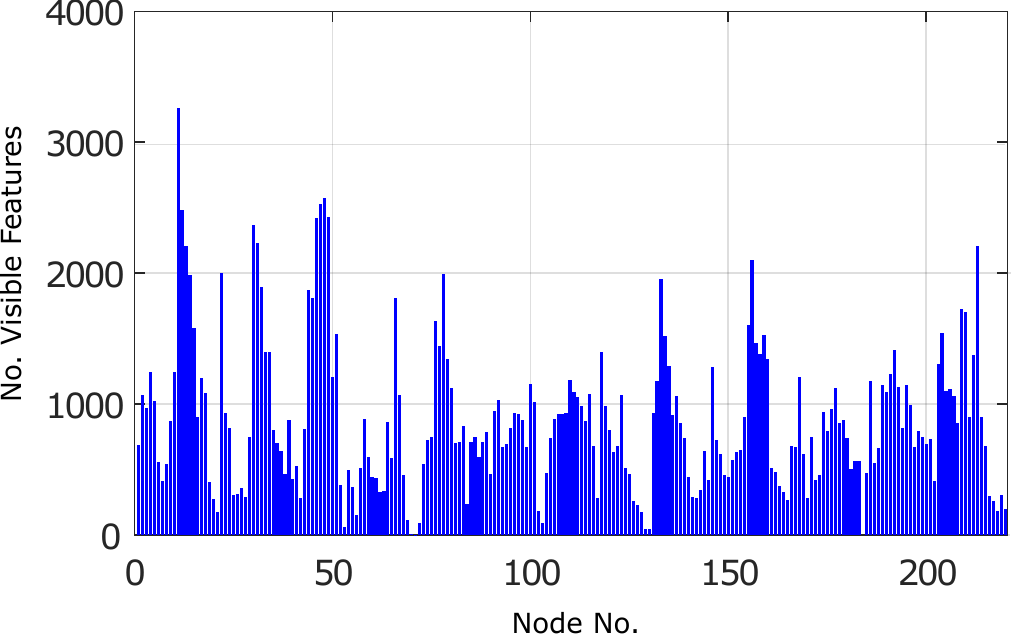}
    \caption{Number of features visible for each node in the museum environment test.}
    \label{fig:no_nodes}
\end{figure}

In terms of efiiciency, a comparison between EM-LOC and the three competing approaches are reported in the right three subplots of Fig. \ref{fig:evaluation}. As is evident from the results of A2, the simple means of using the side-channel information as a prior estimate for selecting nodes in which to localise is not very effective. At low
values of $k$ localisation is attempted in many nodes,
leading to low efficiency and slow performance, while at
high $k$, nodes in which localisation would have been accurate are missed. 
The reason for this can easily be seen from \ref{fig:similarities} where it is evident that the WiFi feature
similarity does not perfectly mirror the true metric similarity.
From the figure it is clear that EM-LOC successfully localises significantly more incoming images than the competing approaches at the same $k$ values. Another noticeable trend is that even at $k=1$, EM-LOC is still able to localise most images, meaning that the localisation can be performed extremely efficiently (every incoming frame only has to be matched against the 3D features contained by a single node). The localisation efficiency of EM-LOC is extremely high, and thus
the efficency is significantly higher for EM-LOC even
compared to the A3 where a trained SVM is used to select candidate nodes.

In Fig. \ref{fig:accuracy_timing} we report a comparison and breakdown of the on-line localisation
processing times for EM-LOC and the three competing methods. Due to the efficiency EM-LOC significantly cuts down the processing time spent on 2D-
3D feature matching compared to A1 which uses all nodes in the matching process. 

\begin{table}[h!]
\centering
\caption{Efficiency of the 6-DoF localisation for the 3 methods}
\label{my-label}
\begin{tabular}{l||l|l|l}
Method & k = 10 & k = 5  & k = 1  \\ \hline \hline
EM-LOC & 0.4783 & 0.4534 & 0.4534 \\ \hline
A1     & 0.1043 & 0.0519 & 0.0187 \\ \hline
A3     & 0.1317 & 0.1257 & 0.1078
\end{tabular}
\end{table}

In summary, by localising only against a select subset of nodes we are able to perform 6-DoF pose updates more frequently than exhaustive matching and also avoid occasional visual false positives, suppressing outlier pose estimates and resulting in a higher localization accuracy. Although the side-channel information is an ideeal means of selecting candidate nodes for localisation, the noisy nature of this data means that existing methods such as using independent features (as in A2) or even an SVM that models correlations between features (as in A3) is not able to provide sufficiently accurate candidate nodes. By making use of a stream of side-channel information along with rough odometry estimates the EM-LOC system can significantly boost the accuracy of the node proposals.

%\begin{figure}[h!]
%    \centering
%    \includegraphics[width=\columnwidth]{Images/efficiency_baseline.pdf}
%    \caption{a) Localisation efficiency of the baseline for comparison using
%WiFi. The maximum efficiency for the baseline is $15\%$. b) Plot of $Neff$
%for localsation using 200 particles. Here $\tau_eff$ has been chosen as 170.}
%    \label{fig:sampling}
%\end{figure}

\section{Conclusion}
\label{sec:conclusion}

In this paper, we proposed EM-LOC, an indoor localisation
system which leverages multiple modalities for efficient global localisation.The key challenge here is the extremely high
computational cost incurred when trying to match features to
the millions of 3D features stored on a localisation graph. We
address this by intelligently integrating multiple modalities
in our localisation system, allowing us to significantly reduce
the matching cost and thereby achieve real-time performance.
Unlike other methods, our system is able to utilise a temporal stream of side-channel information in order to select candidate nodes in which to localise. 
The results show that our multi-modal approach not only
increases the localisation accuracy but significantly reduces
computational time compared to other approaches. For future work, we plan to focus on simultaneous, lightweight human and humanoid localisation.

\bibliographystyle{plain}
\bibliography{refs} 

\end{document}